\documentclass[lettersize, journal]{IEEEtran}		

\usepackage{amsmath,amsfonts}
\usepackage{algorithmic}
\usepackage{adjustbox}
\usepackage{caption}
\usepackage{subcaption}
\usepackage{array}
\usepackage[caption=false,font=normalsize,labelfont=sf,textfont=sf]{subfig}
\usepackage{textcomp}
\usepackage{stfloats}
\usepackage{url}
\usepackage{authblk}

\usepackage{algorithm}
\usepackage {algorithmic}

\usepackage{verbatim}
\usepackage{graphicx}
\usepackage{bm} 
\usepackage{float}

\usepackage[table]{xcolor} 
\usepackage{tabularray}
\usepackage{tabularx}
\usepackage{array}      
\usepackage{booktabs}    

\usepackage{cite}

\usepackage{amsmath}

\usepackage{multicol}
\usepackage{multirow}
\usepackage{balance}	

\usepackage[switch]{lineno} 

\def\BibTeX{{\rm B\kern-.05em{\sc i\kern-.025em b}\kern-.08em
    T\kern-.1667em\lower.7ex\hbox{E}\kern-.125emX}}

\newcolumntype{s}{>{\hsize=.33\hsize\linewidth=\hsize}X}
\newcolumntype{D}{>{\hsize=.4\hsize\linewidth=\hsize}X} 

\newcommand{\wdImg}{\dimexpr \linewidth} 

\begin{document}
\title{Reducing hyperparameter sensitivity in measurement-feedback based Ising machines

\thanks{\hrule \vspace{0.25em}  *Correspondence and requests should be addressed to T.S. (email: toon.jan.b.sevenants@vub.be) or to G.V. (email: guy.verschaffelt@vub.be)}}

\author[*1]{Toon Sevenants}
\author[1]{Guy Van der Sande}
\author[1]{Guy Verschaffelt}

\affil[1]{Applied Physics Research Group, Vrije Universiteit Brussel, Brussels, Belgium} 

\maketitle

\section*{Abstract}
Analog Ising machines have been proposed as heuristic hardware solvers for combinatorial optimization problems, with the potential to outperform conventional approaches, provided that their hyperparameters are carefully tuned. Their temporal evolution is often described using time-continuous dynamics. However, most experimental implementations rely on measurement-feedback architectures that operate in a time-discrete manner. We observe that in such setups, the range of effective hyperparameters is substantially smaller than in the envisioned time-continuous analog Ising machine. In this paper, we analyze this discrepancy and discuss its impact on the practical operation of Ising machines. Next, we propose and experimentally verify a method to reduce the sensitivity to hyperparameter selection of these measurement-feedback architectures.

\section{Introduction}\label{sec: Introduction}
\IEEEPARstart{T}{here} exists a wide range of combinatorial optimization problems (COPs) in society \cite{Barahona1988, Lucas2014, Horvath2016, Saad2001, Wang2020}. 
Many of them are classified as NP-hard, meaning that it is believed there exists no efficient algorithm that can solve this type of problem with polynomial resources \cite{Garey2002}. The underlying reason is that these problems involve finding an optimal discrete configuration, defined by a cost function, from a set of possibilities that grows rapidly (e.g., exponentially) with problem size. Traditional digital computers face significant challenges in terms of time and energy consumption when trying to solve COPs \cite{Anand2020, Horowitz2014}. This has prompted a search for alternative computing paradigms, especially in physics-inspired computing devices. This type op devices are often stochastic in nature and trade accuracy for speed- and energy-efficiency \cite{Siddique2015}. \newline
One of these promising devices is the Ising machine (IM), which is expected to efficiently solve COPs, by leveraging the physics of energy minimization. This system has the potential to outperform traditional digital methods in both speed- and energy-efficiency, making them attractive for solving large-scale, complex optimization problems \cite{Mohseni2022}.\newline
The working princple of any IM is based on the possibility to map the cost function that is associated to a COP, to the Ising Hamiltonian, which is given by
\begin{equation}
\label{eq:ising_energy}
    H_{ising} = -\frac{1}{2} \sum_{ij} J_{ij} \sigma_i \sigma_j - \sum_{i} b_i \sigma_i 
\end{equation}
where $\sigma_i \in \{-1, 1\}$ are binary variables, conventionally referred to as spins, $J_{ij}$ is the interconnection matrix specifying which spins are coupled and with what interaction weight, and the values $b_i$ represent the external fields, which in this paper are set to zero. Dedicated hardware implementations of IMs are then meant to emulate this spin nework and return the minimum energy configuration of the Ising model that, due to the mapping, correspond to the solution of the original optimization problem. \newline
There already exist several hardware implementations of IMs, both in the electrical and optical domain \cite{Mohseni2022, Shim2017, Pierangeli2019,  Takemoto2019, Fabian2019, Honjo2021, English2022, Litvinenko2023, Mallick2023}. Some of them direclty implement the binary spins $ \{ \sigma_i \}$ in their hardware, while other implementations use analog spins $ \{ x_i \}$ instead. In the latter system, which we refer to as analog IMs, the Ising energy can still be calcultated by mapping the analog spin value to binary ones, most commonly by using the sign-function: $sign(x_i) = \sigma_i$. An advantage of this choice is that much hardware is already inherently analog, while it also allows for the use of analog optimization methods such as gradient descent or momentum, which are not feasible in the binary domain. Furthermore, analog hardware components come with an inherent noise that is crucial in the proper working of IMs \cite{Pierangeli2019}. \newline
For these reasons, the analog IM holds great promise, which is why we will focus on them in this paper. Note that the use of analog spins can cause heterogeneity in the spin-amplitudes, which decreases the performance of the analog IM. However, recent works have focussed on overcoming this issue \cite{Leleu2019, Inui2022}. \newline
In simulations, analog IMs are typically modelled to be fully analog, implying that their dynamics evolve continuously over time \cite{Fabian2021, Toon2025}. However, at the moment, many analog IM implementations rely on hybrid setups that combine both analog and digital components, where the digital components are almost always used to implement the spin coupling that is dictated by the interconnection matrix \cite{Fabian2019, Honjo2021}. Altough these implementations still use analog spin variables, their dynamics do not operate continuously over time but rather at specific clock samples, i.e. in discrete iterations. 

The starting point of this paper is our experimental observation that the hyperparameters found from the time-continuous analog IM simulations fail to yield solutions on our opto-electronic hybrid analog IM hardware setup \cite{Fabian2019}. To obtain insight, we investigate the underlying dynamical differences between the time-discrete and time-continuous analog IM. Our analysis shows that there is a non-trivial rescaling effect between the two parameter ranges for which the analog IM successfully converges to optimal solution. We first qualitatively discuss these findings, afterwhich we propose and experimentally validate a method to counter the aformentioned reduction in hyperparameter range.

\section{Methods}
\subsection{Time-discrete vs time-continuous analog IMs}\label{sec:Time-discrete analog IMs}
\begin{figure*}[t]
	\centering
		\includegraphics[width= 1\textwidth]{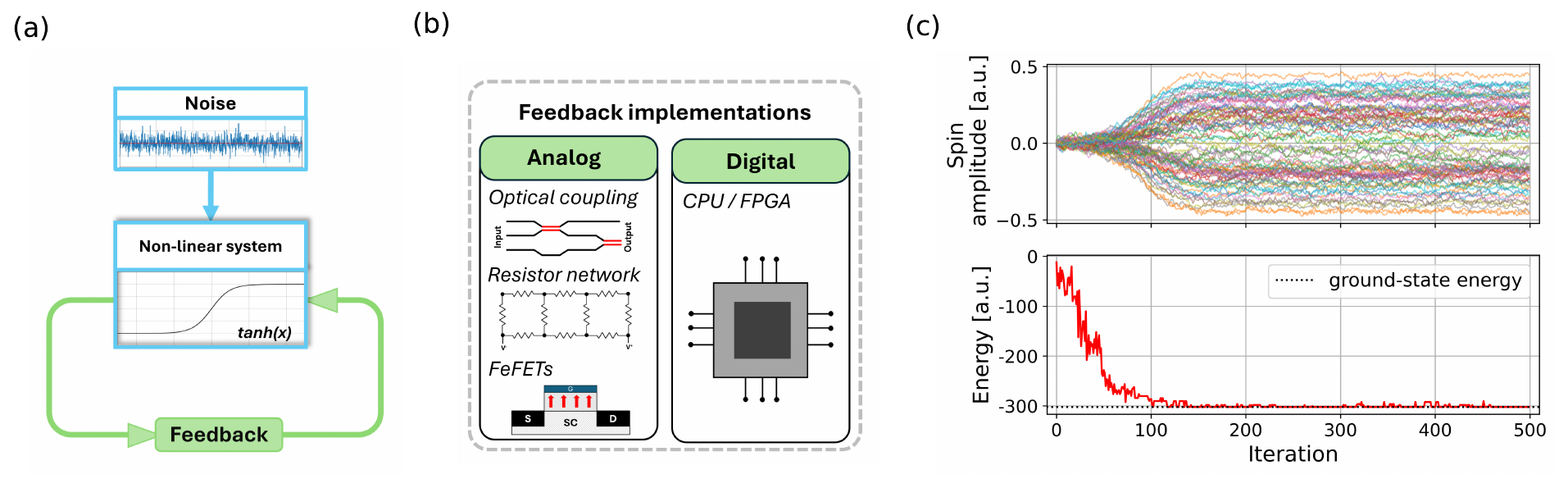}   
		\caption{\textbf{Basic working principle of an analog IM.} \textbf{(a)} Schematic illustration of the operation of an analog IM. Both analog noise, which helps the system escape local minima, and a feedback signal, which drives the system toward lower energy levels, are essential for all analog IM implementations. The feedback signal can be realized in several ways, some of which are shown in \textbf{(b)}. 
		\textbf{(c)} shows an example simulation result of a time-discrete analog IM solving the $g05\_80.1$ MaxCut problem. The upper panel displays the simulated spin amplitudes, while the lower panel shows the corresponding energy evolution. 
}
	\label{fig:IM_schematic_layout}
\end{figure*}

	An analog IM can solve a COP by allowing its spin variables to evolve dynamically over time such that the energy, given by Eq.~\ref{eq:ising_energy}, decreases toward the ground state. A general schematic of the working principle of an analog IM is shown in Fig.~\ref{fig:IM_schematic_layout}(a). This layout holds for both time-continuous and time-discrete analog IMs, which both require a non-linearity, gaussian noise to escape local minima, and a feedback signal to update the values of the analog spin variables $\{x_i\}$. This feedback signal consists of two contributions and is given by
\begin{equation}
      f(t) = \alpha x_i(t) + \beta\sum_j J_{ij}x_j(t)
\label{eq:feedback}
\end{equation}
where $\alpha$ denotes the gain and $\beta$ the coupling strength between spins. The first term represents the gain and reinforces the current state of a spin, while the second term corresponds to mutual spin coupling and drives each spin according to the states of the spins to which it is coupled. As shown in Fig.~\ref{fig:IM_schematic_layout}(b), we distinguish between analog-implemented feedback signals and digitally implemented feedback signals. Although both systems employ this same feedback signal, there is a subtle difference in how the feedback loop is implemented. Both will be explained in more detail in this section.

In a fully analog IM, both the coupling and the feedback signal are implemented directly within the analog hardware itself, for example through a resistor network \cite{Rnetwork_2019}, optical interference \cite{OpticalDelay_2016}, or through CMOS-compatible ferroelectric field-effect transistors \cite{fFET_2023}. As a result, the system continuously evolves according to its internal dynamics toward states of lower energy, without requiring any digital components. Hence, we refer to this system as a time-continuous analog IM. The dynamics of an analog spin variable $x_i(t)$ are then described by a first-order differential equation. Several different nonlinearities can be used to model the specific hardware implementation. In this paper we employ a hyperbolic tangent. Our choice of this nonlinearity is motivated by its ability to capture the clipping effects that arise in hardware implementations due to limited dynamic range and saturation of physical components ~\cite{Fabian2021}. The resulting dynamical equation for spin-$i$ is then given by
\begin{equation}
\label{eq:transfer_function_tc}
      \frac{dx_i}{dt'} = \frac{1}{\tau_\ell} \biggl( -x_i(t') + \tanh\! \bigl( f(t'-\tau_d) + \gamma \zeta_i(t')  \bigr) \biggr)
\end{equation}
where $\tau_d$ represents the feedback-delay, \emph{f(t'-$\tau_d$)} the feedback-term and $\zeta_i$ a stochastic noise term that models the inherent noise of analog hardware, with $\gamma$ being its amplitude. The time $t'$ is expressed in the same units as the relaxation time of the analog hardware $\tau_\ell$, which is determined by the effective bandwidth of the analog system. The linear term $-x_i$ represents dissipation that occurs in a hardware setup, and ensurs that the system relaxes toward a steady state. When considering time-continuous simulations, we take into account that $\tau_d$ is typically much smaller than $\tau_\ell$, and is therefore often omitted from the equation. Since $\tau_\ell$ is determined by the specific analog components used in a given setup, it is more useful to renormalize the equations by introducing the dimensionless time variable $t = \frac{t'}{\tau_\ell}$, which results in
\begin{equation}
\label{eq:transfer_function_tc2}
\frac{dx_i}{dt} = F_i = -x_i(t) + \tanh\! \bigl( f(t) + \gamma \zeta_i(t) \bigr).
\end{equation}
where the time variable $t$ is unitless. We refer to $F_i$ as the transfer function of spin variable $i$. \newline
In a hybrid measurement-feedback setup, on the other hand, the feedback loop no longer consists solely of analog components. Instead, the coupling and feedback signals are processed digitally, for instance on an FPGA, which requires an explicit measurement of the spin states before the calculation can be carried out in the digital domain. The conversion between the analog and digital domains, as well as the digital processing itself, increases the feedback delay $\tau_d$ such that it typically becomes larger as the relaxation time $\tau_\ell$ of the analog subsystem. Therefore, we cannot omit $\tau_d$ from the dynamical equation anymore. A major consequence of this larger delay time is that the analog dynamics settles to a steady state before the digitally processed signal is fed back. In this regime, the spin states are therefore updated iteratively rather than evolving continuously in time, and we refer to such systems as time-discrete analog IMs. Since we measure the spin amplitudes at discrete moments, after they have reached steady state, we rewrite time as
\begin{equation}
t \;=\; k \cdot \frac{\tau_d}{\tau_\ell},
\end{equation}
where we again use time renormalization, and $k (\in \mathbb{N}$) denotes the iteration number. The $\tau_d/\tau_\ell$-ratio is a constant dimensionless factor, allowing the dynamics of time-discrete analog IM to be expressed in terms of unitless iteration number $k$. Using this definition, Eq.~(\ref{eq:transfer_function_tc}) reduces to the following discrete-time map:
\begin{equation}
\label{eq:transfer_function_td}
x_i[k] \;=\; \tanh\!\bigl( f[k-1] + \gamma\,\zeta_i[k] \bigr),
\end{equation}
The feedback term $f[k-1]$ is calculated inside a digital system, e.g. a CPU or FPGA, at discrete time steps using the measured spin amplitudes from the previous iteration. Note that we have omitted the $\tau_d/\tau_\ell$-ratio in Eq.~(\ref{eq:transfer_function_td}), as it only rescales time but does not influence the dynamics.

\subsection{Modeling and benchmarking an analog IMs }\label{sec: SIM_analog IMs}
To simulate an analog IM on a conventional digital computer, we use the dimensionless Eq.~(\ref{eq:transfer_function_tc2}) and (\ref{eq:transfer_function_td}). Eq.~(\ref{eq:transfer_function_td}) can be simulated directly, as it explicitly provides the spin update based on the spin amplitudes at the previous iteration. In contrast, Eq.~(\ref{eq:transfer_function_tc2}) is implemented using the Euler method. The resulting update rule for the spin dynamics is then given by
\begin{equation}
 \label{eq:transfer_function_tc_sim}
    x_i[k] = x_i[k-1] + h\cdot F_i[k-1]
\end{equation}
where $h$ denotes the (Euler) time step, which must be chosen sufficiently small to capture the time-scale of the spins dynamics. In our simulations of a time-continuous analog IM, we take $h = 0.01$.  Remark that to simulate a time-discrete analog IM, we can instead choose $h = 1$, in which case Eq.~(\ref{eq:transfer_function_tc_sim}) reduces to Eq.~(\ref{eq:transfer_function_td}). Since the choice of $h$ is effectively the only difference between the simulation of these two systems, we use it throughout this paper to explicitly distinguish between time-continuous and time-discrete analog IMs. An example simulation of both the spin and energy evolution of a time-discrete analog IM is shown in Fig.~\ref{fig:IM_schematic_layout}(c). The problem being solved is \texttt{g05\_80.1} which is a benchmark problem from the freely accessible BiqMac library. This library contains MaxCut benchmark problems (see Supplementary Note 1) which are frequently used benchmarks for analog IMs as they are proven to be NP-hard \cite{Lucas2014, Garey2002}. Fig.~\ref{fig:IM_schematic_layout}(c) shows that the spin amplitudes \{$x_i$\} are initialized around zero, but due to the feedback signal, these amplitudes grow and bifurcate into either a positive or negative value, corresponding to a spin-up or spin-down state, respectively.

\section{Results and discussion}
\subsection{Shrinking of the hyperparameter range}\label{sec: shrinking_AOO}

\begin{figure}[t]
	\centering
		\includegraphics[width=1\columnwidth]{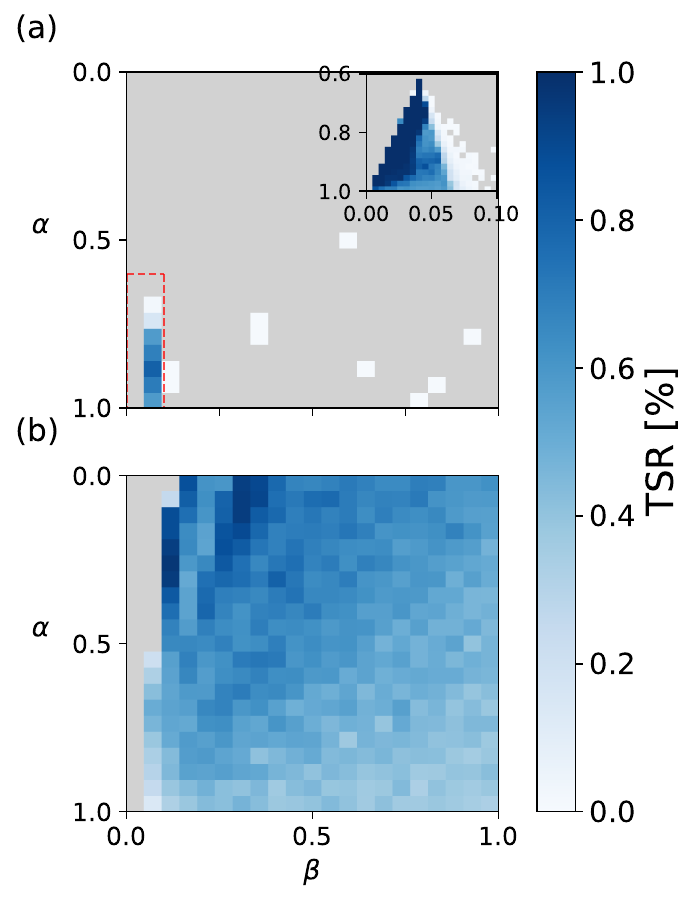} 
		\caption{\textbf{Comparison of the brute force parameter grid scan of a time-discrete and a time-continuous analog IM, for the \texttt{g05$\_$80.1}-benchmark problem.} Both figures have coupling strength $\beta$ and the gain $\alpha$ on the x- and y-axis respectively, and the transient success rate (TSR) color coded. For the time-discrete system shown in \textbf{(a)}, the parameter range leading to non-zero TSR is significantly smaller then for the time-continuous system shown in \textbf{(b)}. The inset in \textbf{(a)} shows a high-resolution  zoom-in of the area indicated by the red dashed box.}
	\label{fig:parameter_space_comparison}
\end{figure}

\begin{table}[t] 
\centering
\caption{\textbf{The AOO, expressed in \%, for the g05-benchmark set from the \emph{biqmac} library}. The AOO of both the time-discrete (h=1) and the time-continuous (h=0.01) simulations are given for each problem. The parameter boundaries of the used scans are $\alpha$ $\in$  [0.5; 1.0] and $\beta$ $\in$ [0.0; 0.5], with a resolution of 21 $\times$ 21. The last row shows the average ratio of the AOO of the time‑continuous analog IM to the AOO of the time‑discrete analog IM for each problem size.}  
\medskip
\begin{tabular}{|c|cc|cc|cc|}
\hline
\multirow{2}{*}{\begin{tabular}{@{}c@{}} \textbf{problem} \\ \textbf{index ($n$)} \end{tabular}}
    & \multicolumn{2}{c|}{\textbf{g05\_60.n}} 
    & \multicolumn{2}{c|}{\textbf{g05\_80.n}} 
    & \multicolumn{2}{c|}{\textbf{g05\_100.n}} \\
\cline{2-7}
    & h=1 & h=0.01 & h=1 & h=0.01 & h=1 & h=0.01 \\
\hline
\textbf{0}  & 13.4 & 90.2 & 4.8  & 86.2 & 2.5 & 84.8 \\
\textbf{1}  & 10.7  & 90.7 & 9.3 & 92.5  & 3.6 & 87.3 \\
\textbf{2}  & 9.3 & 89.6 & 3.2 & 79.6 & 2.7 & 85.9 \\
\textbf{3}  & 4.8  & 71.4 & 3.6 & 85.3 & 2.0 & 79.1 \\
\textbf{4}  & 10.0  & 88.9 & 5.0 & 86.8 & 1.1 & 79.6 \\
\textbf{5}  & 9.5 & 88.7 & 2.3 & 77.6 & 2.0 & 78.5 \\
\textbf{6}  & 3.2 & 62.8 & 3.4 & 84.1 & 2.9 & 87.3 \\
\textbf{7}  & 11.8  & 87.3 & 2.5 & 78.9 & 2.3 & 83.7 \\
\textbf{8}  & 6.6 & 81.2 & 2.5 & 73.0 & 1.1 & 70.7   \\
\textbf{9}  & 9.1 & 87.3 & 4.3 & 85.7 & 2.9  & 85.3 \\
\hline
Average ratio  & \multicolumn{2}{c|}{10.7} & \multicolumn{2}{c|}{23.3} & \multicolumn{2}{c|}{40.1} \\

\hline
\end{tabular}
\label{tab:AOO_biqmac_values}
\end{table}

A characteristic feature of heuristic solvers such as the IM is that they do not always find the optimal solution of the problem at hand in a single run. In this work, we quantify the solver's performance using the percentage of successful runs. Specifically, we use the transient success rate (TSR), which is defined as the probability that the analog IM finds the solution during a single run. Another characteristic of heuristic solvers is that they require a set of hyperparameters that must be optimized for the solver to work efficiently. Analog IMs also exhibits this behavior and have two main hyperparameters that control their dynamics: the gain $\alpha$ and the mutual coupling strength $\beta$. To identify combinations of $(\alpha,\beta)$ that allow the analog IM to find the solution of a given problem, we perform a brute-force parameter scan. An example of such a scan is shown in Fig.~\ref{fig:parameter_space_comparison}. In this approach, the system is evaluated over a predefined set of hyperparameter combinations, and a specific performance metric, in this case the TSR, is mapped onto a blue color scale. The hyperparameter combination with the highest TSR is then considered to be the optimal hyperparameter setting. In this paper, the TSR is always calculated using 250 runs with 5000 iterations per run, unless explicitly stated otherwise. \newline
Fig.~\ref{fig:parameter_space_comparison} shows the parameter scan for the biqmac \texttt{g05\_80.1} benchmark problem, for both a time-discrete and a time-continuous analog IM. Altough both scans use identical parameter boundaries of $[0,1]$ for $\alpha$ and $\beta$, a clear difference can be observed between them.
The parameter scan of the time-discrete analog IM, shown in Fig.~\ref{fig:parameter_space_comparison}(a), shows a much smaller region with a non-zero TSR compared to that of the time-continuous analog IM, shown in Fig.~\ref{fig:parameter_space_comparison}(b). The inset in Fig.~\ref{fig:parameter_space_comparison}(a) provides a zoomed-in view of the region indicated by the red dashed box.  Comparring Fig.~\ref{fig:parameter_space_comparison}(a) with (b), it is clear that for the time-discrete system the range of good hyperparameters has shrunk compared to the time-continuous case. \newline
To quantify this reduction, we define the area of operation (AOO) as the percentage of hyperparameter combinations in a parameter scan that result in a non-zero TSR. This metric allows for a direct comparison of the range of useful hyperparameter combinations between different parameter scans, provided that the same problem, parameter boundaries and scan resolution are used. The AOO values for all \texttt{g05} benchmark problems, for both time-discrete and time-continuous systems, are listed in Tab.~\ref{tab:AOO_biqmac_values}. For this analysis, the parameter boundaries are chosen as $\alpha \in [0.5;1.0]$ and $\beta \in [0.0;0.5]$, with a scan resolution of $21 \times 21$. The three columns with results in Tab.~\ref{tab:AOO_biqmac_values} correspond to problems with 60, 80 and 100 spins respectively. From Tab.~\ref{tab:AOO_biqmac_values}, we observe an order of magnitude difference in the range of useful hyperparameter combinations between the time-discrete and time-continuous cases for all considered benchmark problems. The AOO of the time‑discrete system is at least ten times lower than that of the time‑continuous system, while for larger problems it is on average even forty times lower.\newline
To qualitatively understand why this shrinking occurs, we look back at the difference between the time-discrete and the time-continous analog IMs. From Sec.~$\ref{sec: SIM_analog IMs}$, we observe that the only difference between the two systems is the used time-step $h$. Writing Eq.~\ref{eq:transfer_function_tc_sim} explicitly gives us
\begin{equation}
 \label{eq:transfer_function_tc_sim_full}
    x_i[k] = (1-h) \cdot x_i[k-1] + h\cdot \text{tanh}( f[k-1] )
\end{equation}
From this expression, it is evident that decreasing $h$ increases the contribution of the linear term while simultaneously reducing the influence of the tangent hyperbolic.  As a consequence, the effective impact of the hyperparameters $\alpha$ and $\beta$ is substantially reduced when reducing $h$, since they determine the feedback signal $f$ inside the tangent hyperbolic. The next spin state is thus mainly determined by the previous one. From a physical perspective, this corresponds to a system that both updates its spin state in small increments and also rapidly adjust its feedback. This enables the system to update its spins more smoothly and navigate the complex energy landscape more effectively by avoiding large overshoots. 

\begin{figure*}[t]
    \begin{center}
        \begin{minipage}[t]{1\textwidth}
            \centering
            \includegraphics[width=\wdImg]{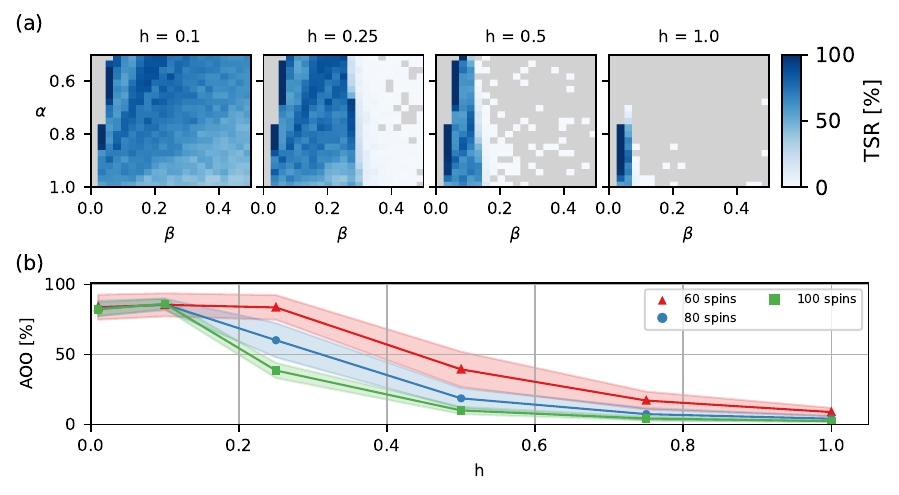} 
        \end{minipage}
    \end{center}
    \caption{\textbf{Influence of the h-value on the parameter space.} 
    			\textbf{a} The brute force parameter grid scan of the g05$\_$80.1 problem for four different h-values. The x- and y-axes of the grid scans are the coupling strength $\beta$ and the gain $\alpha$, respectively. The color bar indicates the transient success rate (TSR), while a zero-valued TSR is represented by the grey color.
     			\textbf{b} The area of operation (AOO), defined as the percentage of ($\alpha$,$\beta$)-combinations that results in a non-zero transient success rate (TSR), as a function of the h-value for different problem sizes. Each problem size has a separate symbol and color. The color shaded areas represent the different problem instances of a specific problem size.}
    \label{fig:h_influence}
\end{figure*}

\noindent In contrast, in the time-discrete case with $h=1$, the next state is determined entirely by the nonlinear map of Eq.~\ref{eq:transfer_function_td}. As a result, the change after one iteration in the spins' amplitude becomes larger, meaning that the system performs larger updates between iterations compared to the time-continuous analog IM. Simply decreasing the gain or coupling strength is not effective, since the feedback signal must remain sufficiently strong to induce bifurcation. The signal should therefore neither be too small nor too large, ensuring that bifurcation occurs while still allowing the spins to properly navigate the energy landscape. This leads to a significantly more delicate balance between hyperparameters and ultimately results in a much narrower AOO, as observed for the time-discrete analog IM. Physically, this corresponds to a system that updates its spin states strongly and relatively slowly in time. \newline
Next, we emphasize that this shrinking in the hyperparameter range is not just a trivial rescaling. If the effect of the time step could be captured by a simple multiplicative factor to adjust $\alpha$ and $\beta$, the parameter ranges of the time-discrete and time-continuous formulations would be directly related. To demonstrate that this is not the case, we examine in more detail how the time step $h$ interacts with the hyperparameters. Specifically, we expand Eq.~\ref{eq:transfer_function_tc_sim_full} in a Taylor series up to third order in the weak-coupling regime, $\alpha \gg \beta$. This expansion allows us to disentangle the influence of the time step on the individual hyperparameters~\cite{Fabian2021}. For a $h \neq$ 1, Eq.~$\ref{eq:transfer_function_tc_sim_full}$ then becomes:

\begin{equation}
\label{eq:h_rescaling2}
\begin{split}
    x_i[k] &= \Big (1 - h + h\alpha \Big) x_i[k-1] - h\frac{\alpha^3 x_i^3[k-1]}{3} \\
    & + h\beta \sum_{j} J_{ij} x_j[k-1] \\ 
\end{split}
\end{equation}
while for $h$=1, the equation becomes
\begin{equation}
\label{eq:h_rescaling3}
\begin{split}
    x_i[k] &= \alpha x_i[k-1] - \frac{\alpha^3 x_i^3[k-1]}{3} + \beta \sum_{j} J_{ij} x_j[k-1] \\ 
\end{split}
\end{equation}
where we have negelected the noise term in both equations for clarity. \newline
The Taylor expansion allows us to identify that there is non-trivial difference between the dynamics of the time-continuous and the time-discrete analog IMs. When going from \( h \)=\( 1 \) to a \( h \) $\neq$ \( 1 \) we have:

\begin{equation}
\left\{
\begin{aligned}
   & \alpha^3 \rightarrow h \alpha^3 \\
   & \beta \rightarrow h\beta \\
      & \alpha \rightarrow (1-h + h\alpha) \\
\end{aligned}
\right.
\end{equation}
From this expression, we observe that the first two terms scale proportional to $h$, whereas the last term exhibits a different transition with $h$. This shows that there is not a simple rescaling factor to convert the optimal hyperparameters for the time-continuous analog IM to those for the time-discrete analog IM.

\subsection{Increasing the hyperparameter range}
\begin{figure}[t!]
	\centering
	\includegraphics[width=1.0\columnwidth]{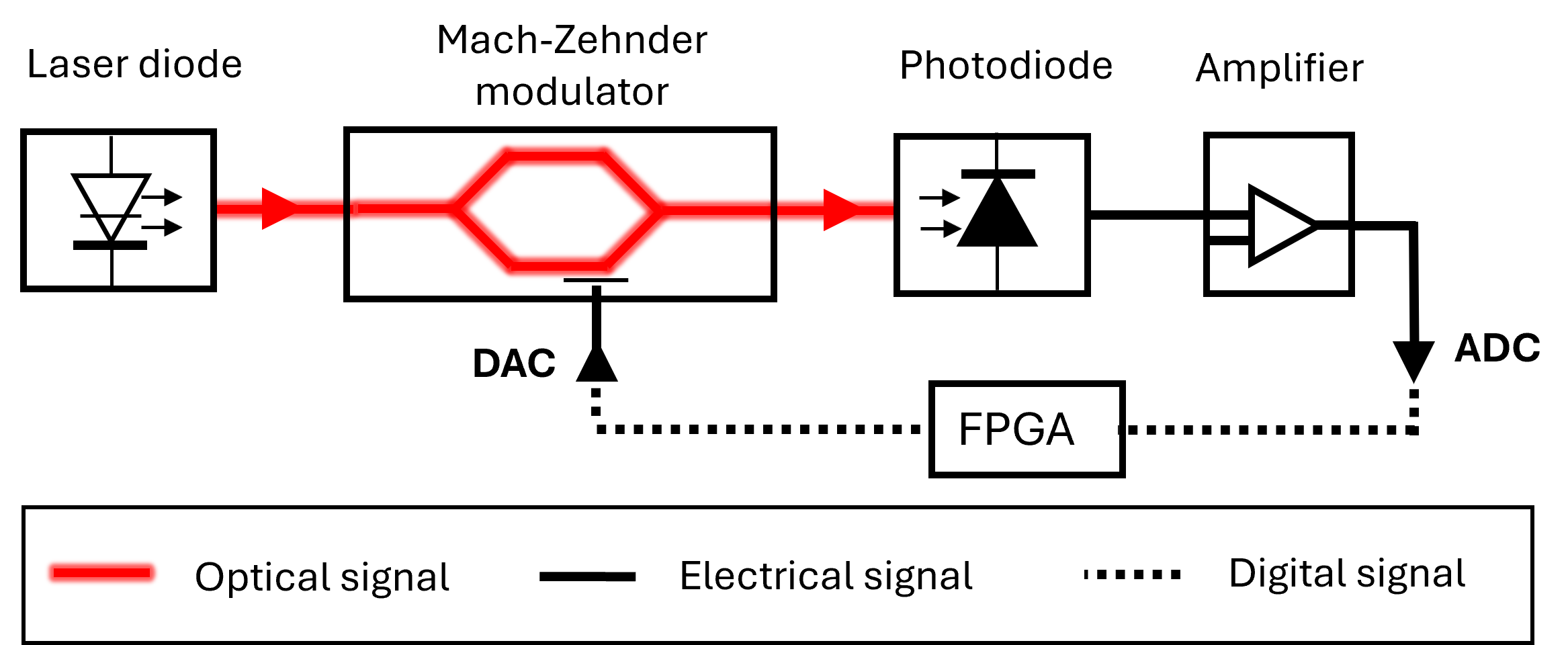}
	\caption{\textbf{Schematic of the time-multiplexed opto-electronic CIM}. In this hybrid setup, the input signal is optically modulated, then converted into an analog voltage and finally sampled, digitized, and stored on an FPGA. The calculated feedback signal is converted back into a voltage and sent to the modulation input of the MZM.}
	\label{fig:oeo-setup}
\end{figure}
From the discussion of Eq.~\ref{eq:h_rescaling2} in the previous section, we expect that the transition from small to large $h$ values is not characterized by a sharp cutoff or threshold, but rather by a gradual shift in the AOO. For $h<1$, the update rule remains a weighted combination of the previous spin state and the applied feedback, such that changes in $h$ continuously modify the relative contribution of these terms. As a result, changing $h$ should lead to a gradual change in the system dynamics rather than an abrupt transition. We therefore propose this tuning of the value of $h$ as a method to mitigate the shrinking of the AOO in the time-discrete analog IM, and demonstrate that it is effective not only in numerical simulations but also in hardware. Note that we have explored several alternative approaches to influence the AOO, including long runtimes, variations in noise strength, and the spin-sign method~\cite{DePrins2025}. However, none of these approaches had a measurable or consistent influence on the AOO (see Supplementary Note 2). \newline
In simulations, we calculate the AOO for different values of $h$ between the two extreme cases of $h=1$ and $h=0.01$. Although choosing an arbitrary value of $h$ is non-conformal in the sense that it does not correspond to a specific physical time-continuous hardware implementation, it is nevertheless algorithmically allowed. Moreover, $h \neq 1$ can be incorporated in the time-discrete system simply when calculating the feedback signal digitally in the measurement-feedback loop. Fig.~\ref{fig:h_influence}(a) shows the parameter scans of the \texttt{g05\_80.1} benchmark problem for four different values of $h$. As $h$ increases from $0.01$ to $1.0$, we observe a reduction in the hyperparameter range that yields a non-zero TSR. 

\begin{figure*}[b!]
    \begin{center}
        \begin{minipage}[b]{1\textwidth}
            \centering
             \includegraphics[width=\wdImg]{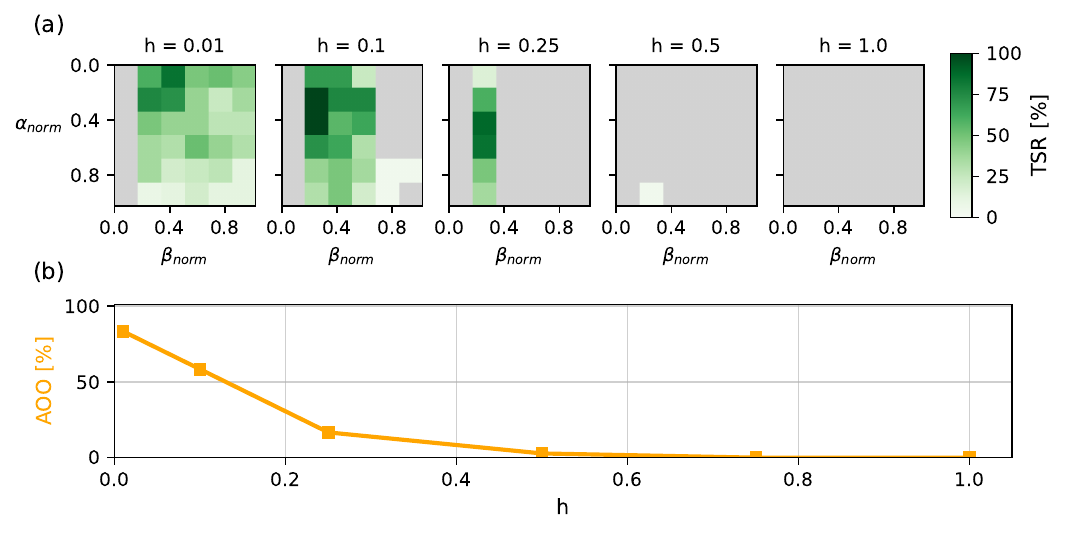}
        \end{minipage}
    \end{center}
    \caption{\textbf{Experimental verification of the decreasing parameter sensitivity with decreasing $h$-value.}  \textbf{(a)} Parameter scans over the normalized \( \alpha_{\mathrm{norm}}, \beta_{\mathrm{norm}} \)-parameter space for the \texttt{g05\_80.1} benchmark problem, shown for five different values of \( h \).  The evaluated metric is the transient success rate (TSR).  As \( h \) decreases, a larger region of the hyperparameter space yields a non-zero TSR. \textbf{(b)} Resulting AOO values corresponding to the six evaluated \( h \)-values are indicated by symbols (line is intended to guide the eye).
}
    \label{fig:exp_h_influence}
\end{figure*}

To quantify this shrinkage, we again use the AOO and plot the corresponding values for all \texttt{g05} benchmark problems in Fig.~\ref{fig:h_influence}(b), using in total six different values of $h$ lying between $0.01$ and $1.0$. The data points represent the average AOO for each problem size, calculated over 10 benchmarks per problem size, and the shaded area around them represents the standard deviation of this distribution. For a fair comparison, we still calculate the TSR for each $(\alpha,\beta)$ combination and each value of $h$ using 5000 iterations per run and a total of 250 runs. \newline
For all considered problems, the AOO decreases significantly as $h$ approaches 1, reducing the viable hyperparameter range. As a result, the time-discrete analog IM becomes substantially more sensitive to hyperparameter choices compared to the time-continuous analog IM. Moreover, the data suggest a size-dependent effect, meaning that this increased sensitivity becomes even more pronounced for larger problem instances. At the same time, the figure shows that the change in AOO is not abrupt. Decreasing the value of $h$ gradually increases the AOO in a smooth and continuous manner. From Fig.~\ref{fig:h_influence}(b), we observe that this behavior is consistent across all benchmark problems, highlighting the potential of tuning $h$ to adjust the hyperparameter sensitivty in time-discrete analog IMs. Note that the AOO values appear to saturate for the smallest $h$ values. We attribute this saturation to the fixed parameter boundaries, implying that although the true parameter space likely continues to expand beyond these limits, this growth cannot be captured within the imposed bounds. However, decreasing $h$ also reduces the error in the Euler integration of Eq.~\ref{eq:transfer_function_tc_sim}. Once $h$ is sufficiently small, the dynamics closely approximate those of the time-continuous analog IM, and further reductions in $h$ will therefore not to lead to changes in the useful hyperparameter ranges.

\subsection{Experimental verification}
To further support our claims, we validate the simulation results on a hardware setup. For this purpose, we make use of the opto-electronic “poor man’s Coherent Ising machine”, of which a schematic outine is shown in Fig.~(\ref{fig:oeo-setup}) \cite{Fabian2019}. This setup consists of a standard DFB laser diode ($\lambda$ = 1.55$\mu\text{m}$), a Mach–Zehnder modulator (MZM), a photodiode, and a field-programmable gate array (FPGA) equipped with 14-bit ADC and DAC components. At the start, the spins are encoded sequentially as optical amplitudes in a pulse train propagating through an optical fiber, i.e., they are time-multiplexed. These optical spins are read out by a photodiode and converted into electrical voltages, which are subsequently digitized by the ADC and stored on the FPGA. \newline
Once all spins have been measured, the FPGA computes the feedback signal according to Eq.~\ref{eq:feedback} using a coupling matrix \( \mathbf{J} \) that is also loaded onto the same FPGA. Finally, the feedback signal for each spin is sent from the FPGA to the DAC, converted into an analog electrical signal, and applied to the MZM, where it modulates the optical amplitude of the spins for the next iteration. This time-multiplexed measurement-feedback setup operates as a time-discrete system, allowing it to be modeled using Eq.~(\ref{eq:transfer_function_td}). We can also use this measurement-feedback setup to investigate the effect of different values of $h$, thus effectively implementing Eq.~(\ref{eq:transfer_function_tc_sim}). To do this, we modified the software running on the FPGA, in particular the scaling of the feedback signal $f_i$ by multiplying it with the value of $h$. A pseudo-code of this implementation is provided in Supplementary Note~3. \newline
Before performing the measurements, we carry out a gauging procedure in the form of a bifurcation scan. In this scan, the experimental gain \( \alpha_e \) is increased from zero untill the spins start to bifurcate. We perform this gauging step because there is no direct one-to-one mapping between the hyperparameters used in simulation and those of the physical experimental system. Since theory predicts that the bifurcation occurs at \( \alpha = 1 \), we can identify the corresponding experimental gain at which this transition occurs. In our setup, this value is \( \alpha^{\mathrm{bif}}_{\mathrm{exp}} = 2.34 \), which we subsequently use as the conversion factor between the hyperparameters of the simulation and experiment (see Supplementary Note~4 for more details).\newline
To experimentally observe the gradual transition predicted by Fig.~\ref{fig:h_influence}, we perform a brute-force grid scan over $\alpha_{\mathrm{exp}}$ and $\beta_{\mathrm{exp}}$ on the \texttt{g05\_80.1} benchmark problem. This problem was chosen because it exhibits a high average TSR value in simulation for both \( h \) = \( 1 \) and \( h \) = \( 0.01 \). 
In total, we perform a hyperparameter scan of the experimental setup for six different values of \( h \), ranging from 1 down to 0.01. To balance resolution and measurement time, we restrict the scan to a \(6\times 6\) grid in the \((\alpha_{\mathrm{exp}}, \beta_{\mathrm{exp}})\)-plane, performing 25 runs for each parameter combination. Each run contains 2000 iterations, independent of the chosen \( h \)-value. The experimental results are shown in Fig.~\ref{fig:exp_h_influence}(a), where the TSR is mapped onto a green color scale. Both axes are normalized by the experimental bifurcation value, i.e. $\alpha_{\mathrm{norm}} = \alpha_{\mathrm{exp}} / \alpha^{\mathrm{bif}}_{\mathrm{exp}}$ and $\beta_{\mathrm{norm}} = \beta_{\mathrm{exp}} / \alpha^{\mathrm{bif}}_{\mathrm{exp}}$. This normalization ensures that the resulting hyperparameters \( \alpha_{\mathrm{norm}} \) and \( \beta_{\mathrm{norm}} \) both lie within the interval \([0,\,1]\), like the simulation results shown in Fig.~\ref{fig:parameter_space_comparison}. \newline
The outermost right panel in Fig.~\ref{fig:exp_h_influence}(a) ,  showing the results for \( h = 1 \), corresponds to the standard operating conditions of the time‑discrete analog IM. No (\(\alpha_{\mathrm{norm}}, \beta_{\mathrm{norm}}\))-combination within the considered parameter range and resolution give a non‑zero TSR. This highlights the hyperparameter fine‑tuning required for this type of experimental setup. As \( h \) decreases, multiple parameter combinations begin to appear for which the TSR becomes non‑zero, confirming the overall trend observed in simulations. This trend is also captured in the associated AOO- of these parameter scans shown in Fig.~\ref{fig:exp_h_influence}(b). A particularlly usefull hyperparameter range appears for $h \leq 0.25$. \newline
The transition observed in the experimental results is not as smooth as in the simulations. This is most likely due to the limited resolution of the experimental parameter scan, as higher resolutions are not easily accessible because of time constraints. As a result, the use of lower $h$ values is even more important in experiments, as they increase the likelihood of accessing regions with a non-zero TSR within a coarse parameter scan. 

\section{Discussion}
We experimentally confirmed that the use of a small artificial Euler step $h$ increases the hyperparameter range in a time‑discrete measurement–feedback setup, making the system less sensitive to parameter initialization and tuning. Moreover, since the optimal hyperparameters are obtained through simulations, a reduced $h$-step also makes the setup considerably less sensitive to errors in the simulation‑to‑hardware parameter conversion. \newline
Finally, in this work we use the hyperbolic tangent as the system's nonlinearity \cite{Fabian2021}. However, the results presented here are independent of the particular choice of nonlinear function. Other implementations employing different nonlinearities, such as the cosine-squared or third-order nonlinearity, encounter similar difficulties in identifying useful hyperparameters. This is because the underlying problem is not the specific nonlinearity itself, but rather the time-discrete measurement--feedback dynamics. Consequently, all time-discrete analog IMs benefit from the insights developed in this paper.

\section{Conclusion}

In this paper, we have investigated the difference between the time-continuous and time-discrete dynamics of analog IMs. The latter implementation is observed to have a much smaller range of useful parameter values, making it more sensitive to the hyperparameter tuning. This poses a problem since many of the today's analog IM implementations are hybrid ones that follow the time-discrete dynamics. To address this limitation, we have proposed and experimentally verified that the introduction of an small artificial Euler integration step $h$ allows to reduce the sensitivity to parameter initialization, tuning and conversion. Moreover, this solution applies to any hardware platform because it is independent of the chosen nonlinearity. Future research will aim to understand why the AOO seems to decrease with increasing problem size, and how the introduced $h$-step affects the overall time‑to‑solution of analog IMs.

\section{Data availability} \label{sec:  Data availability}
\noindent The authors declare that all relevant data are included in the manuscript. Additional data are available from the corresponding author upon reasonable request.

\section{Code availability} \label{sec:  Code availability}
\noindent The used software code is available from the corresponding author upon reasonable request.

\section{Author contributions} \label{sec: Author contributions}
\noindent T.S. performed the simulations and the experiments and wrote the manuscript. G.V.d.S and G.V. supervised the project. All authors discussed the results and reviewed the manuscript.

\section{Additional information} \label{sec:  Additional information}
\noindent \textbf{Competing interest: } G.V.d.S. is an Editorial Board Member for Communications Physics but was not involved in the editorial review of, or the decision to publish this article. All the other authors declare no competing interests. \newline
 	
\noindent  \textbf{Acknowledgements: } This research was funded by the Research Foundation Flanders (FWO) under grant G0A6L25N. Additional funding was provided by the EOS project ”Photonic Ising Machines”. This project (EOS number 40007536) has received funding from the FWO and F.R.S.-FNRS under the Excellence of Science (EOS) programme.


\onecolumn

\title{SUPPLEMENTARY NOTES}




\maketitle

\section*{\textbf{SUPPLEMENTARY NOTE 1: MAX-CUT PROBLEM}}

Solving maximum cut (MaxCut) optimization problems is often used to evaluate the performance of Ising machines. The MaxCut problem is known to be NP-hard, meaning that finding the optimal solution is considered computationally difficult. This difficulty is reflected in recent publications reporting new best-known solutions, which demonstrate that the problem remains challenging even for state-of-the-art methods \cite{NewBestCut_2025_1}. \newline

The MaxCut problem on an undirected graph \(G=(V,E)\) is defined as as the task of splitting the vertices of the graph into two sets so as to maximize the total weight of edges that run between these two sets. We denote this total weight, i.e., the cut value, by \(C\). Let $w_{ij}$ be the symmetric coupling matrix of the graph, with $w_{ii}=0$ and $w_{ij}\ge 0$. \newline
To describe a cut we attach a spin \(s_i\in\{+1,-1\}\) to each vertex \(i\). Vertices with the same spin value lie on the same side of the cut. An edge \((i,j)\) is counted in the cut exactly when \(s_i\neq s_j\). The weighted cut value is then given by
\begin{equation}
C \;=\; \sum_{i\neq j} w_{ij}\,\mathbf{1}_{\{s_i\neq s_j\}} .
\end{equation}
We can rewrite the indicator \(\mathbf{1}_{\{s_i\neq s_j\}}\) in terms of spins:
\begin{equation}
\mathbf{1}_{\{s_i\neq s_j\}} \;=\; \frac{1 - s_i s_j}{2},
\end{equation}
because \(s_i s_j = +1\) when the vertices are on the same side and \(-1\) when they are on opposite sides. Substituting this into the expression for \(C\) and simplifying gives
\begin{equation}
C \;=\; \frac{1}{2}\sum_{i,j} w_{ij}\,\frac{1 - s_i s_j}{2}
\;=\; \frac{1}{4}\sum_{i,j} w_{ij} \;-\; \frac{1}{4}\sum_{i,j} w_{ij}\,s_i s_j .
\end{equation}
Where the extra \(1/2\) in front removes the double counting coming from the summation. Now bringing the first term on the right to the left yields
\begin{equation}
	\label{eq:MaxCut_1}
-\frac{1}{4}\sum_{i\neq j} w_{ij}\,s_i s_j \;=\; C \;-\; \frac{1}{4}\sum_{i\neq j} w_{ij}.
\end{equation}
If we define the Ising energy as
\begin{equation}
H(\{s_i\}) \;=\; -\frac{1}{2}\sum_{i,j} J_{ij}\,s_i s_j,
\end{equation}
where \(J_{ij} = -w_{ij}\), we can rewrite Eq.~\ref{eq:MaxCut_1} as
\begin{equation}
H(\{s_i\}) \;=\; - \left( 2C \;+\; \frac{1}{2}\sum_{i,j} J_{ij} \right),
\end{equation}
where the second term on the right-hand side is a constant, independent of the spin configuration \(\{s_i\}\). From this relation, we see that maximizing the cut value \(C\) is equivalent to minimizing the Ising energy \(H(\{s_i\})\). \newline

In this paper, we use the ``\texttt{g05\_}'' benchmark set from the freely accessible BiqMac library. All instances in this set are MaxCut problems with 60, 80, or 100 spin variables. For each problem size, we use 10 instances. For these relatively small MaxCut instances, the exact solutions are known and documented in Ref.~\cite{biqmac_library}.

\newpage
\section*{\textbf{SUPPLEMENTARY NOTE 2: INVESTIGATING OTHER INFLUENCES }}

In addition to the influence of the time step $h$, we also evaluated several other factors that might reduce the parameter sensitivity of the time-discrete analog IM. Specifically, we examined three possible approaches: increasing the runtime, adjusting the noise strength, and applying the spin-sign method \cite{DePrins2025}. As in the previous experiments, we used the three problem sizes from the ``\texttt{g05\_}'' benchmark library (60, 80, and 100 spins) and included all ten instances available for each problem size. For all three approaches, the parameter scan was conducted over $\alpha \in [0.5, 1.0]$ and $\beta \in [0.0, 0.5]$ using a $21 \times 21$ resolution grid. For each $(\alpha, \beta)$ pair, the transient success rate (TSR) was determined from 250 independent runs. The number of iterations and the noise levels used in the experiments are specified separately for each method in the following sections.

\subsection{Runtime influence}
\begin{figure}[H]
\centering
\includegraphics[width=1\columnwidth]{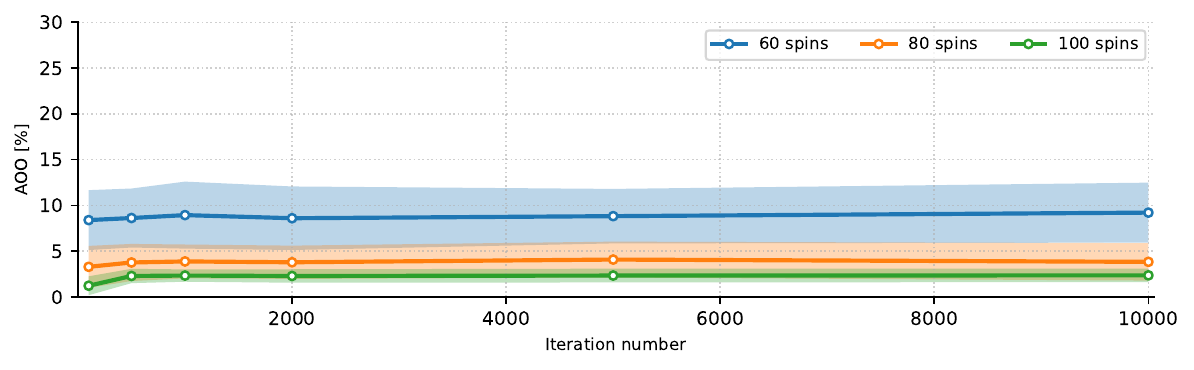} 
\caption{
\textbf{The influence of the runtime on the AOO for three different problem sizes.} In total, we used 6 different runtimes, ranging from 100 to 10 000 iterations.
}
\label{fig:RuntimeInfluence_AOO}
\end{figure}

To investigate the effect of the runtime, we evaluated six different values (expressed in iterations): 100, 500, 1000, 2000, 5000, and 10\,000 iterations. For all runtime settings, we used a fixed noise strength of $\gamma = 0.01$. In addition, we considered three problem sizes, as described earlier. The resulting AOO values are shown in Fig.~\ref{fig:RuntimeInfluence_AOO}. The shaded regions around the data points indicate the variation in AOO across the ten instances available for each problem size. As we can see from this figure, there isn't any significant difference in AOO when changing the runtime over two orders of magnitude. We set the runtime to 5000 iterations in this paper for simulations and 2000 iterations in the experiment. 

\subsection{Noise influence}

\begin{figure}[H]
\centering
\includegraphics[width=1\columnwidth]{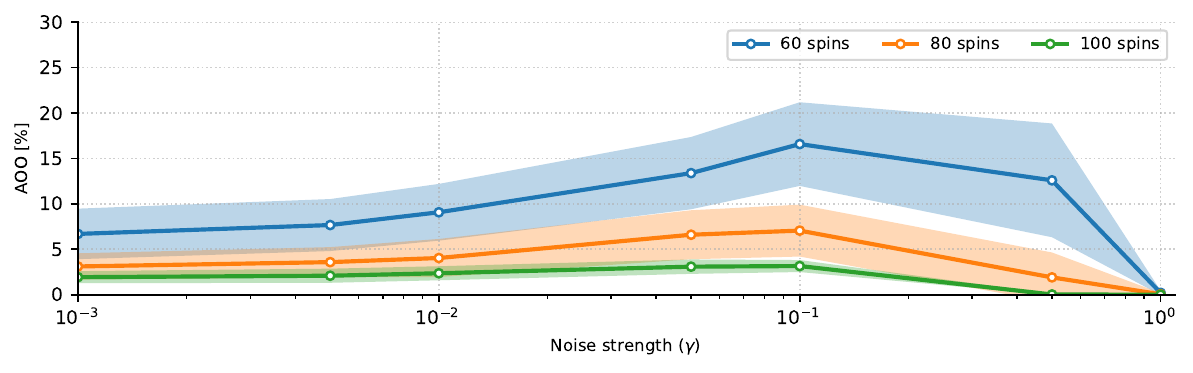} 
\caption{
\textbf{The influence of the noise on the AOO for three different problem sizes.} In total, 7 different noise strengths were used, ranging from 0.001 to 1.0.
}
\label{fig:Noiseinfluence_AOO}
\end{figure}

To investigate the effect of the noise strength, we evaluated seven different values: 0.001, 0.005, 0.01, 0.05, 0.1, 0.5, and 1.0. For all the noise settings, we used a fixed runtime of 5000 iterations. Also here, we considered the three problem sizes from the ``\texttt{g05\_}'' benchmark library. The resulting AOO values are shown in Fig.~\ref{fig:Noiseinfluence_AOO} where the shaded regions around the data points indicate the variation in AOO across the ten instances available for each problem size. Note that the x-axis has a logaritmic scale.

As the noise incrases, we observe a small increase in AOO as well, before dropping to zero for $\gamma$ = 1.0. This increase is mainly visible for the 60-spin problem set, and is almost not observable anymore for the 100-spin set. Moreover, even for the 60-spin problem set the change is relatively small consdering that the AOO for the time-continuous analog IM (for the same parameter scan range and resolution) is found to be around 80\%. We set the noise strength to 0.01 in this paper for simulations. 

\subsection{Spin-sign method}
The spin-sign method was introduced in Ref.~\cite{DePrins2025} to improve the performance of analog IMs on problems that include external fields. In this approach, the binary spin value is used when computing the mutual coupling feedback. Consequently, the feedback signal for spin $x_i$ is modified such that the update equation becomes
\begin{equation}
    f_i = \alpha x_i + \beta \sum_{j} J_{ij}\,\sigma(x_j),
\end{equation}
where $\sigma(x_j)$ denotes the sign of $x_j$. This substitution ensures that the interaction term more accurately reflects the discrete nature of the Ising spins. The underlying idea is that the feedback should be determined by the structure of the Ising energy landscape, rather than by the continuous state variables when these have not yet converged to their binary values. \newline

\begin{figure}[H]
\centering
\includegraphics[width=1\columnwidth]{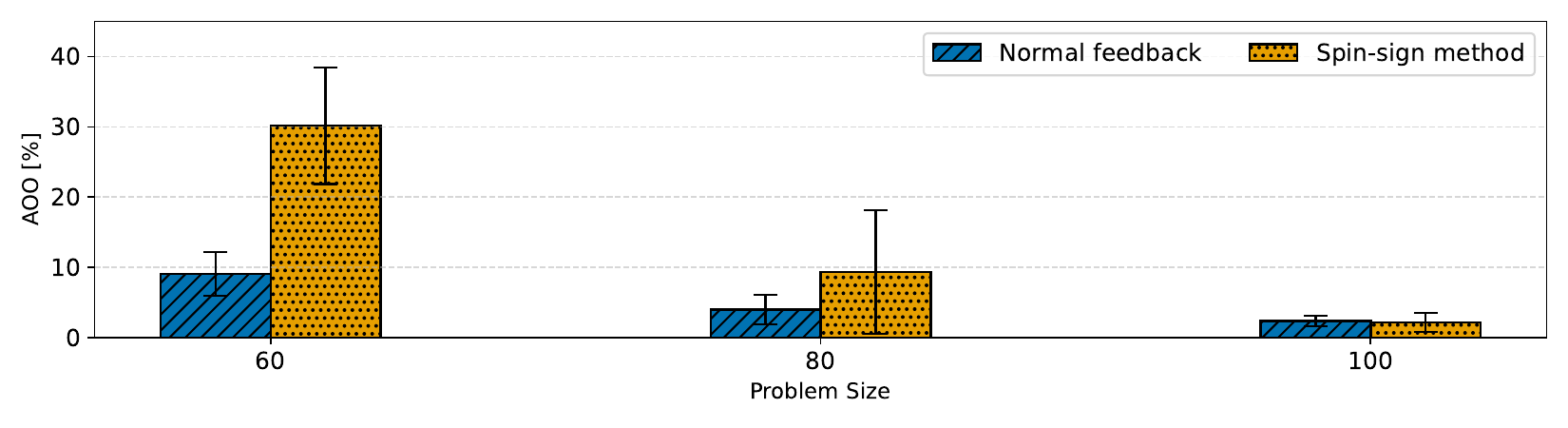} 
\caption{
\textbf{The influence of the spin-sign method on the AOO with a noise strength of $\gamma$=0.01.} The dark blue bars represent the reference simulations performed without the spin-sign method, while the yellow bars correspond to simulations using the spin-sign implementation. All simulations use identical parameters except for the inclusion of the spin-sign method.
}
\label{fig:spinsigninfluence_AOO_1}
\end{figure}

In this work, we do not include external fields, but our goal is to investigate whether this modification enables the analog IM to reach the correct solution for a larger portion of the $(\alpha,\beta)$-parameter space compared to a system without the spin-sign method. The resulting average AOO values for each problem size are shown in Fig.~\ref{fig:spinsigninfluence_AOO_1}, where the dark yellow color bars correpond to the AOO calculated from the simulations using the spin-sign method, and the dark blue color bars corresponds to using a feedback signal based on the spin amplitudes \{$x_i$\}. The error bars indicate the variation in AOO across the ten instances available for each problem size. The used noise strenght and iteration number are respectively 0.01 and 5000 iterations.

From Fig.~\ref{fig:spinsigninfluence_AOO_1} we observe that, for the smallest problem size of 60spins, the AOO shows a noticeable dependence on the method used to calculate the feedback signal. However, the magnitude of this effect remains relatively small compared to the AOO seen in the time-continuous analog IM, which reach values around 80\% over the same parameter-scan range and resolution. Moreover, the influence vanishes with increasing problem size. For the 80-spin instances, the average value is still bigger compared to the average value from the reference system but the error bars shows that there is a wide difference among the problem instances, hence there is no consistent effect. For the 100-spin instances, no significant differences in average AOO are observed. \newline

\begin{figure}[H]
\centering
\includegraphics[width=1\columnwidth]{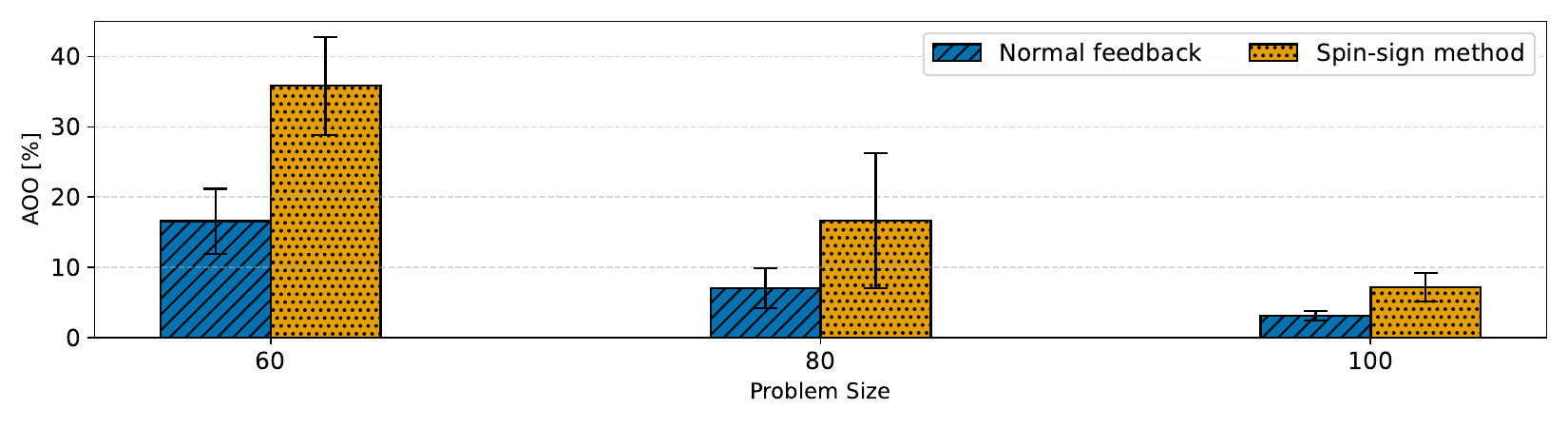} 
\caption{\textbf{Influence of the spin-sign method on the AOO for a noise strength of $\gamma$ = 0.1.} The dark blue bars represent the reference simulations performed without the spin-sign method, while the yellow bars correspond to simulations using the spin-sign implementation. All simulations use identical parameters except for the inclusion of the spin-sign method.
}
\label{fig:spinsigninfluence_AOO_2}
\end{figure}

\begin{figure}[H]
\centering
\includegraphics[width=1\columnwidth]{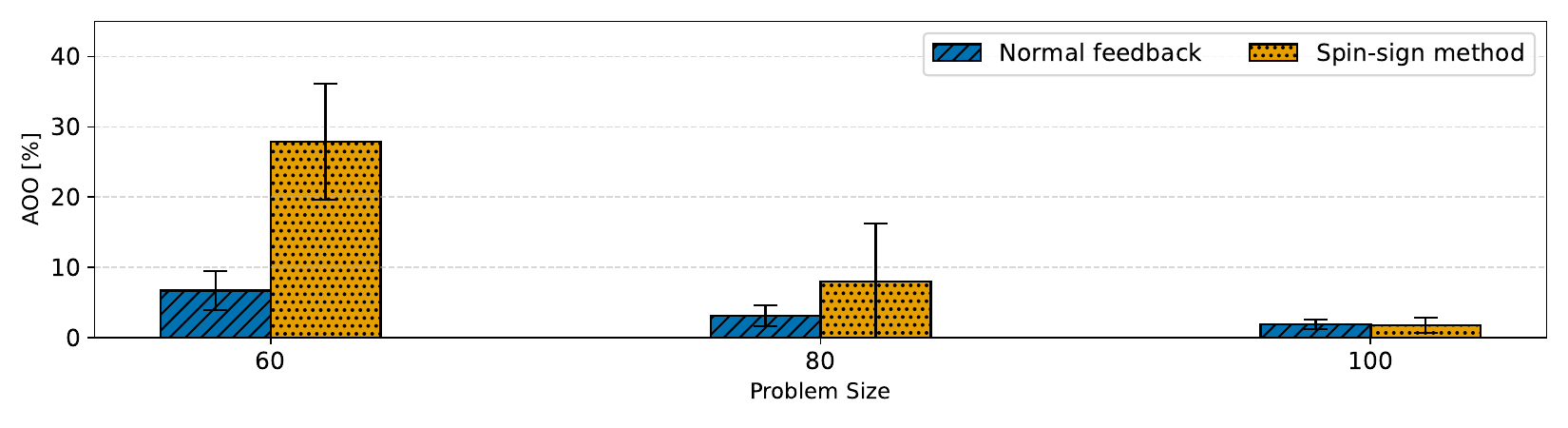} 
\caption{
\textbf{The influence of the spin-sign method on the AOO with a noise strength of $\gamma$=0.001.} The dark blue bars represent the reference simulations performed without the spin-sign method, while the yellow bars correspond to simulations using the spin-sign implementation. All simulations use identical parameters except for the inclusion of the spin-sign method.
}
\label{fig:spinsigninfluence_AOO_3}
\end{figure}

For completeness, we also evaluated the noise dependence of the spin-sign method, as shown in Fig.~\ref{fig:spinsigninfluence_AOO_2} and Fig.~\ref{fig:spinsigninfluence_AOO_3}. Here as well, we do not find any significant increase in AOO across the different problem sizes.

\newpage
\section*{\textbf{SUPPLEMENTARY NOTE 3: PSEUDO-CODE USED IN THE FPGA OF THE EXPERIMENTAL SETUP}}

\begin{algorithm}
\caption{Main analog IM Execution Loop}\label{alg:main}
\begin{algorithmic}[1]

\STATE \textbf{Input:} $N$, noise strength $\gamma$, parameters $a,b,h$, length $T$
\STATE $N \gets N + 10$ \quad (extra trigger and check spins)
\STATE Initialize hardware
\STATE Read matrix $J$ and bias vector $b_i$
\STATE Initialize both $spins \gets 0$, $spins\_old \gets 0$
\STATE Create empty result matrix $R$
\STATE
\FOR{$t = 1$ to $T$}

    \STATE $spins\_new \gets$ sample$(fback)$
    \STATE
    \STATE $spins \gets spins\_old + h(-spins\_old + spins\_new)$
    \STATE $R(:,t) \gets spins$
    \STATE $spins\_old \gets spins$
    \STATE $fback \gets -a \cdot spins - b \cdot J \cdot spins - b_i + \gamma \cdot$ GaussianNoise()

\ENDFOR
\STATE
\STATE Write $R$ to file

\end{algorithmic}
\end{algorithm}

\newpage
\section*{\textbf{SUPPLEMENTARY NOTE 4: BIFURCATION}}

Bifurcation plays a crucial role in the operation of many Ising machine implementations, as it determines the point at which the trivial zero solution becomes unstable and the spins transition into a binary state. This mechanism is also important in the context of the simulation-to-experiment parameter conversion, as discussed in more detail below. \newline

From bifurcation theory, the standard form of a supercritical pitchfork bifurcation is given by
\begin{equation}
    \frac{dx}{dt} = (\alpha - 1)x - x^3,
    \label{eq:standard_bifurcation}
\end{equation}
where the $-x$ term, commonly used in the Ising-machine literature, accounts for the extra loss present in hardware implementations and $\alpha$ is the gain applied to the Ising machine. The pitchfork bifurcation described by Eq.~\ref{eq:standard_bifurcation} occurs, in the absence of spin coupling at $\alpha = 1$. In hardware, however, the 
gain in the experimental setup corresponding to this bifurcation point i snot straightforward to know a priori. This mismatch arises from physical losses that are stronger than the linear loss assumed in theory as well as from digitization effects. The spin amplitudes are namely represented by discrete ADC/DAC values rather than continuous, bounded variables. These hardware-specific influences lead to parameter values that differ from those in the numerical model. \newline

To relate simulation results to experimental behavior, we perform a bifurcation scan, in which the final steady-state spin amplitudes are measured as a function of the gain parameter $\alpha$, without spin coupling (i.e. $\beta$ = 0). This allows us to identify the experimental bifurcation point, denoted $\alpha^{\text{bif}}_{\text{exp}}$, at which the zero-solution loses stability. Since we know from theory that the bifurcation point is $\alpha^{\text{bif}}_{\text{sim}} = 1$,  $\alpha^{\text{bif}}_{\text{exp}}$ effectively serves as a conversion factor that maps the optimal parameters found in simulation to those expected to work in experiment.

\begin{figure}[H]
\centering
\includegraphics[width=1\columnwidth]{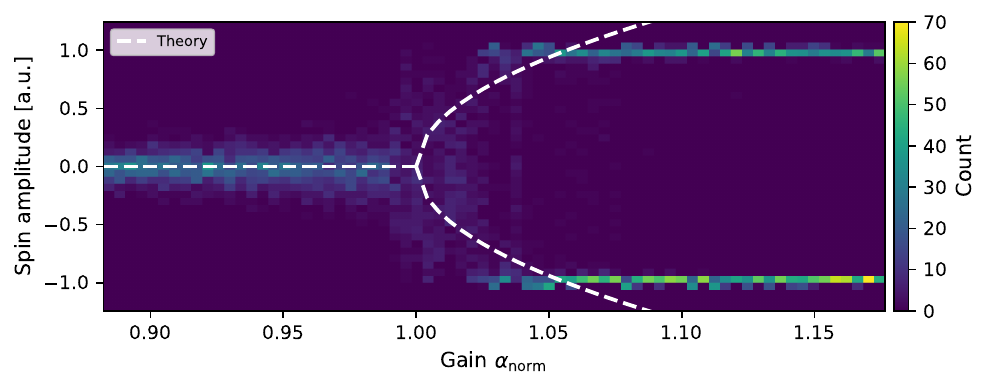} 
\caption{\textbf{Behavior of uncoupled artificial spins around the pitchfork bifurcation.} The figure shows a normalized pitchfork bifrucation. The saturation in the spin-amplitude at higher $\alpha_{norm}$ is due to the clipping effects in the FPGA’s DAC/ADC.
}
\label{fig:exp_bifurcation}
\end{figure}

Fig.~\ref{fig:exp_bifurcation} shows a bifurcation, measured on the OEO-IM \cite{Fabian2019}. For each experimental value of $\alpha$, the uncoupled spins are allowed to evolve freely over 500 iterations. After this time, the final spin states are recorded and plotted as a colored histogram along the y-axis. Repeating this procedure for all experimental $\alpha$ values yields the pitchfork form shown in Fig.~\ref{fig:exp_bifurcation}. In this figure, the spin amplitudes were normalized to lie within the interval $[-1, 1]$, and the experimental $\alpha_{\text{exp}}$-values were rescaled by a factor of $1/2.34 \approx 0.42$ so that the bifurcation occurs at $\alpha_{\text{norm}}$ = $1.0$. This implies that the optimal parameters obtained from simulation must be multiplied by a convertion factor of $\alpha^{\text{bif}}_{\text{exp}}$= $2.34$ to match the experimental setting. On top of the experimentally determined bifurcation, we also plot the theoretical pitchfork bifurcation, described by Eq.~\ref{eq:standard_bifurcation}, in white. Note that for higher values of $\alpha_{\mathrm{norm}}$, the experimentally measured spin amplitudes exhibit saturation. This is due to clipping effects in the FPGA’s DAC/ADC. \newline

It is important to note that, in practice, determining the precise experimental bifurcation point is non-trivial. An exact transition would require infinite runtime of the experimemt, and noise can suppress the spin amplitudes close to bifurcation, masking the true bifurcation point. The combination of the system’s parameter sensitivity and the uncertainty introduced by the conversion procedure descibed above, highlights the difficulty in making hybrid analog IM perform efficiently.

\end{document}